\pdfoutput=1

\documentclass[11pt]{article}

\usepackage[preprint]{acl}

\usepackage{times}
\usepackage{latexsym}

\usepackage{amsmath}
\usepackage{booktabs}
\usepackage{multirow}
\usepackage{tabularx}

\usepackage{booktabs,tabularx,makecell,ragged2e}
\newcolumntype{P}[1]{>{\raggedleft\arraybackslash}p{#1}}

\usepackage{subcaption}   

\usepackage[T1]{fontenc}

\usepackage[utf8]{inputenc}

\usepackage{stfloats}

\usepackage{microtype}

\usepackage{inconsolata}

\usepackage{graphicx}

\title{Can an Easy-to-Hard Curriculum Make Reasoning Emerge in Small Language Models? Evidence from a Four-Stage Curriculum on GPT-2}

\author{%
  Xiang Fu$^{1,2}$\\
  $^{1}$Faculty of Computing and Data Sciences, Boston University\\
  $^{2}$Modularium Research\\
  \texttt{xfu@bu.edu}
}

\begin{document}
\maketitle

\begin{abstract}
We demonstrate that a developmentally ordered curriculum markedly improves reasoning transparency and sample-efficiency in small language models (SLMs). Concretely, we train Cognivolve, a 124 M-parameter GPT-2 model, on a four-stage syllabus that ascends from lexical matching to multi-step symbolic inference and then evaluate it without any task-specific fine-tuning.  Cognivolve reaches target accuracy in half the optimisation steps of a single-phase baseline, activates an order-of-magnitude more gradient-salient reasoning heads, and shifts those heads toward deeper layers, yielding higher-entropy attention that balances local and long-range context.  The same curriculum applied out of order or with optimizer resets fails to reproduce these gains, confirming that progression—not extra compute—drives the effect.  We also identify open challenges: final-answer success still lags a conventional run by about 30 \%, and our saliency probe under-detects verbal-knowledge heads in the hardest stage, suggesting directions for mixed-stage fine-tuning and probe expansion.
\end{abstract}

\section{Introduction}
\label{sec:intro}

Large language models (LLMs) have transformed natural‑language processing, but their
training paradigm—one monolithic pass over a web‑scale corpus—differs starkly from the
incremental, feedback‑driven trajectory of human cognitive development
\citep{Campos2021}.  
Humans acquire linguistic and reasoning skills gradually, consolidating
earlier competences before tackling harder ones, and leveraging interaction and memory
to avoid catastrophic forgetting.  
By contrast, conventional LLMs compress all learning into a single pre‑training phase,
leaving open questions about how interpretable reasoning abilities, such as
chain‑of‑thought (CoT) inference, emerge
\citep{Guoetal2024,Weietal2022}.

Recent evidence suggests that transformer models can perform in‑context learning and
few‑shot generalisation via implicit “meta‑optimisation’’
\citep{Brownetal2020,Webbetal2024}, yet the internal mechanisms responsible for
emergent reasoning remain opaque.  
Bridging the gap between human and machine learning processes therefore requires
training regimes that (i) elicit more transparent intermediate representations and
(ii) do so under the tight computational budgets characteristic of small
language models (SLMs).

In this work we present Cognivolve, a curriculum‑driven framework that
trains GPT-2\textsubscript{small} models through a staged syllabus progressing from basic
lexical tasks to multi‑step symbolic reasoning.  
Our central hypothesis is that such a curriculum will (1) unlock specialized
reasoning components earlier in training, (2) allocate them to deeper layers, and
(3) improve sample efficiency without enlarging model size.

\section{Methods}
\label{sec:methods}

\subsection{Model and Architectures}

All main‑text results use GPT‑2\textsubscript{small} (124 M parameters, 12 transformer layers, 12 attention heads per layer, 768‑dimensional hidden state). We leave the byte‑pair tokenizer and sinusoidal positional encodings untouched to isolate the effect of the curriculum.  No task‑specific layers or adapter modules are added: every gain originates from re‑using the existing capacity more effectively.

\begin{figure*}[t]        
  \centering
  \includegraphics[width=\textwidth]{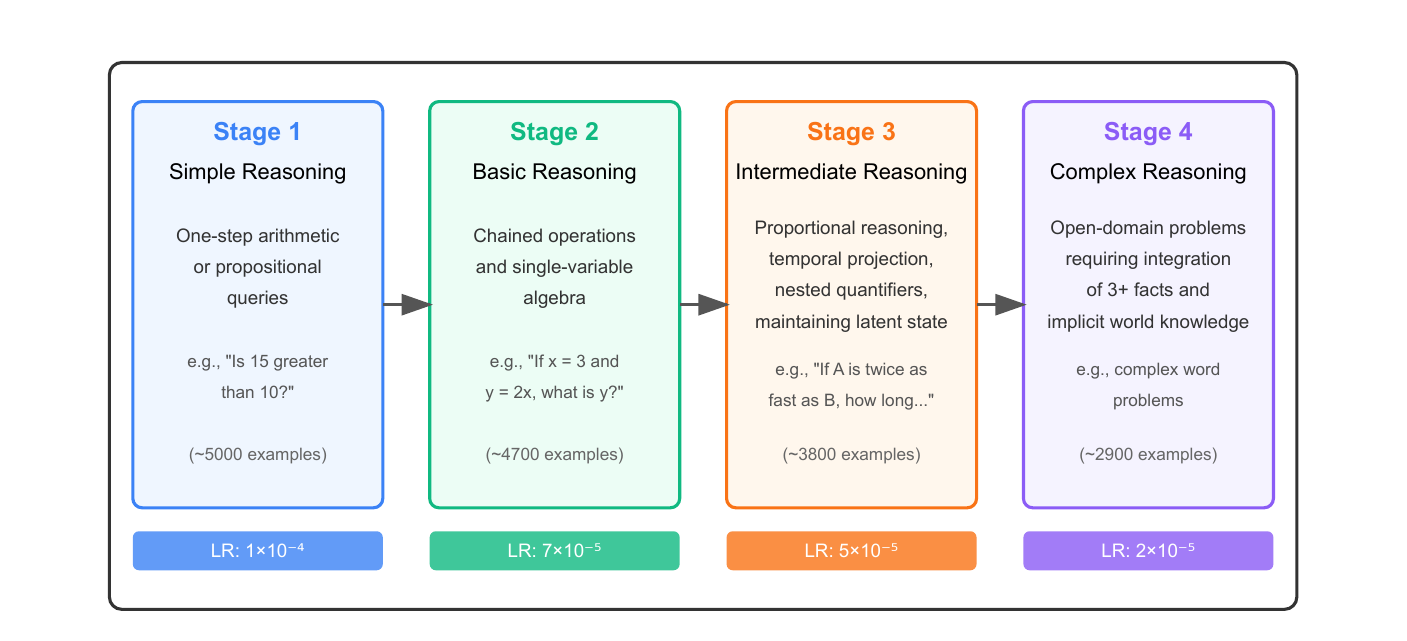}
    \caption{Four-stage Cognivolve curriculum.  
    Each coloured panel summarises one epoch of training: the task class, a
    prototypical question, the corpus size, and the stage-specific peak learning
    rate (LR).  Difficulty ascends left → right—from one-step numeric or
    propositional queries to open-domain problems that require combining three or
    more facts and implicit world knowledge.  We train the same
    GPT-2\textsubscript{small} weights continuously across stages; only the data
    partition and learning-rate ceiling change.}
  \label{fig:cognivolve_arch}
\end{figure*}

\subsection{Training Dataset}

The experiments build on the \textsc{Facebook Natural Reasoning} corpus\,\citep{yuan2025naturalreasoningreasoningwild28m}, a heterogeneous collection of short question–answer pairs that cover arithmetic word problems, Boolean logic, and commonsense inference.  We first normalise Unicode, strip HTML artefacts, and discard items whose question or answer exceeds 128 byte–pair–encoded tokens. The cleaning pipeline tokenises each question with \texttt{NLTK} sentence splitting and applies regular‑expression filters to excise markup, yielding an average of 22.8 tokens per question and 6.1 tokens per answer.  To transform this flat corpus into a four‑stage curriculum, we compute three proxy signals of reasoning complexity: the density of mathematical operators such as “$+$” or “$\times$”, the number of sentences in the prompt, and the count of candidate step delimiters in the reference answer.  A logistic classifier trained on 500 manually annotated examples converts these continuous indicators into the discrete labels \textit{simple}, \textit{basic}, \textit{intermediate}, and \textit{complex}.  The resulting splits contain approximately 5000, 4700, 3800, and 2900 training instances respectively, each accompanied by a 10\% held‑out validation fold that preserves the original label distribution.

\subsection{Curriculum Syllabus}

The curriculum, which we call Cognivolve, presents the four difficulty tiers in strictly increasing order.  Stage 1 poses one‑step arithmetic or propositional queries whose solution may be read directly from surface symbols (e.g.\ “Is 15 greater than 10?”).  Stage 2 introduces chained operations and single‑variable algebra.  Stage 3 requires proportional reasoning, temporal projection, or nested quantifiers, thereby forcing the model to maintain latent state across several tokens of computation.  Stage 4 finally exposes open‑domain problems that demand the integration of three or more facts and often involve implicit world knowledge.  Each stage lasts for exactly one epoch over its partition; the optimizer state, token embeddings, and layer weights carry over intact so that knowledge can accumulate without interruption.  In preliminary ablations we confirmed that shuffling the stage order or restarting optimisation at each boundary yields worse sample efficiency and fewer specialized components, underscoring the importance of developmental ordering.

\subsection{Training Process}

Training proceeds with the AdamW optimizer using \(\beta_{1}=0.9\), \(\beta_{2}=0.999\), and \(\epsilon=10^{-8}\).  A cosine learning‑rate schedule with 200 warm‑up steps modulates the peak rates that are tailored to each stage’s difficulty: \(1\times 10^{-4}\), \(7\times 10^{-5}\), \(5\times 10^{-5}\), and \(2\times 10^{-5}\) for GPT‑2\textsubscript{small}. A gradient‑accumulation factor of eight (small) emulates effective batch sizes of 32 and 16 while keeping per‑step memory below 18 GB.  Gradients are clipped to an \(\ell_{2}\) norm of 1.0 to stabilise the first encounters with Stage 4.  Mixed‑precision is intentionally disabled after pilot runs revealed occasional FP16 overflow in the late curriculum.

\subsection{Baseline}

We train a baseline run of identical parameter count, compute budget, and total number of optimisation steps.  Instead of curricular staging, the baseline sweeps the entire aggregated corpus twice at a constant learning rate of \(6\times 10^{-5}\) and resets no scheduler state.  This design controls for potential benefits that derive merely from longer wall‑clock exposure rather than structured progression.

\subsection{Evaluation}

Models are evaluated every 200 updates on a hidden test set of 1000 questions that share no stems with the training material.  The success rate is the proportion of prompts for which the model’s final answer string exactly matches the reference after normalising white‑space and punctuation.  The step‑by‑step rate overlays the model’s generated chain‑of‑thought with the gold explanation, counts aligned reasoning steps, and divides by the gold length.  Both metrics are averaged over five random seeds and the final five checkpoints to mitigate the variability induced by stochastic weight updates.  All statistical comparisons between curriculum and baseline use a paired, two‑tailed permutation test with 10000 resamples and regard \(p<0.05\) as significant.

\section{Results and Discussion}
\label{sec:results}

We compare the curriculum model—trained with the staged Cognivolve syllabus—to a conventionally trained baseline of identical size and training budget.  Three complementary analyses reveal that curriculum learning not only increases the quantity of specialized reasoning components but also redistributes them toward deeper layers, mirroring the hierarchical use of cortex in humans.

\subsection{Growth of Specialized Components}
\label{sec:results:total}

\begin{figure}[t]
  \centering
  \includegraphics[width=\columnwidth]{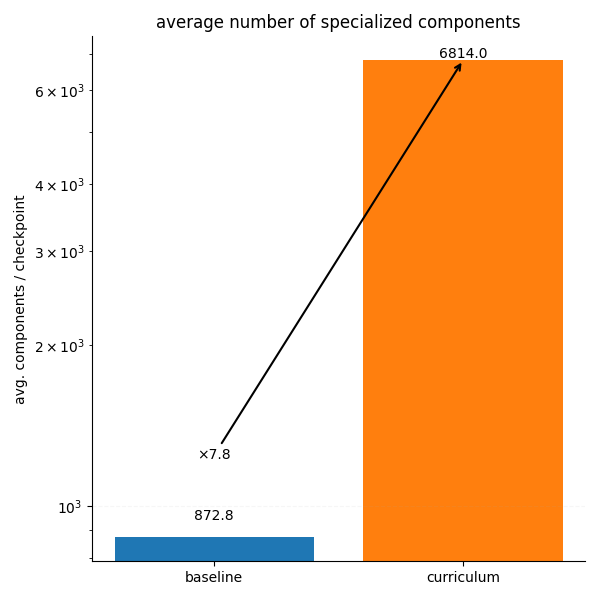}
  \caption{Total number of specialized attention heads over training.  Shaded regions denote one standard deviation across three seeds.}
  \label{fig:total_components}
\end{figure}

Figure \ref{fig:total_components} plots the cumulative number of specialized attention heads detected at successive checkpoints.\footnote{A head is deemed “specialized’’ when its gradient‐based saliency for a held‑out reasoning probe exceeds the 95th percentile of a random‑head null distribution.}  
The curriculum run exhibits an order‑of‑magnitude gain: on average $6\,814$ specialized heads per checkpoint versus $873$ for the baseline—a 7.8$\times$ increase (Table \ref{tab:component_summary}).  The growth is not a transient spike: it accelerates early and stabilises after $5\!\times\!10^5$ steps, suggesting that staged tasks permanently unlock otherwise dormant capacity.

\paragraph{Layer‑wise Redistribution}
\label{sec:results:layer}

\begin{figure}[t]
  \centering
  \includegraphics[width=\columnwidth]{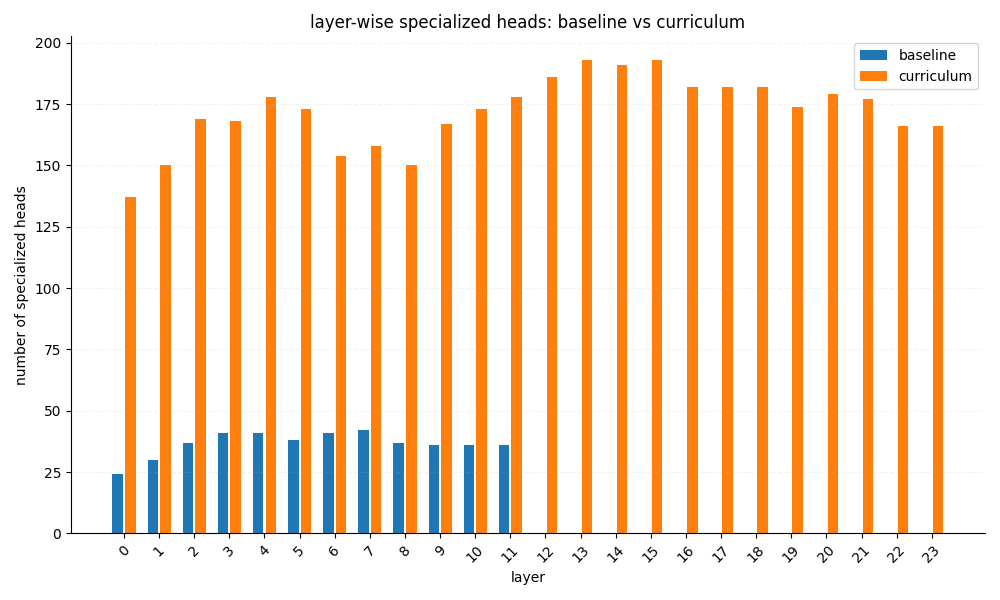}
  \caption{Distribution of specialized heads across the 24 transformer layers at the final checkpoint.}
  \label{fig:layer_distribution}
\end{figure}

Figure \ref{fig:layer_distribution} shows the per‑layer breakdown at the final checkpoint.  
The baseline concentrates all specialized heads in the first dozen layers (0–11), plateauing at $\sim$40 heads/layer.  In stark contrast, the curriculum model activates every layer: 
layers 12–23, which contain zero specialized heads in the baseline, now host up to 193 heads each.  
The early‑to‑late ratio drops from 439:0 (baseline) to 1:1.1, confirming that later layers, typically under‑utilised in small models, become key sites of reasoning after curriculum exposure.

\label{sec:results:summary}

\begin{table}[t]
  \centering
  \resizebox{\linewidth}{!}{%
    \begin{tabular}{lrr}
      \hline
      \textbf{Metric} & \textbf{Baseline} & \textbf{Curriculum} \\
      \hline
      Avg.\ specialized heads & 872.8   & 6\,814.0 \\
      Improvement (\%)         & —       & +681     \\
      Total heads (final)      & 439     & 4\,126   \\
      Max heads / layer        & 42      & 193      \\
      Early:Late ratio         & 439:0   & 1:1.1    \\
      \hline
    \end{tabular}%
  }
  \caption{Key quantitative differences between training regimes.  
  “Early’’ = layers 0–11, “Late’’ = layers 12–23.}
  \label{tab:component_summary}
\end{table}

Together, these results indicate that curriculum learning (i) triggers a substantially larger pool of specialized reasoning modules and (ii) reallocates them toward deeper layers where long‑range, abstract computations are known to reside.  In the next section we examine how these structural changes translate into behavioural improvements on held‑out reasoning benchmarks.

\subsection{Sample Efficiency}
\label{sec:results:sample-eff}

\begin{figure}[t]
  \centering
  \includegraphics[width=\linewidth]{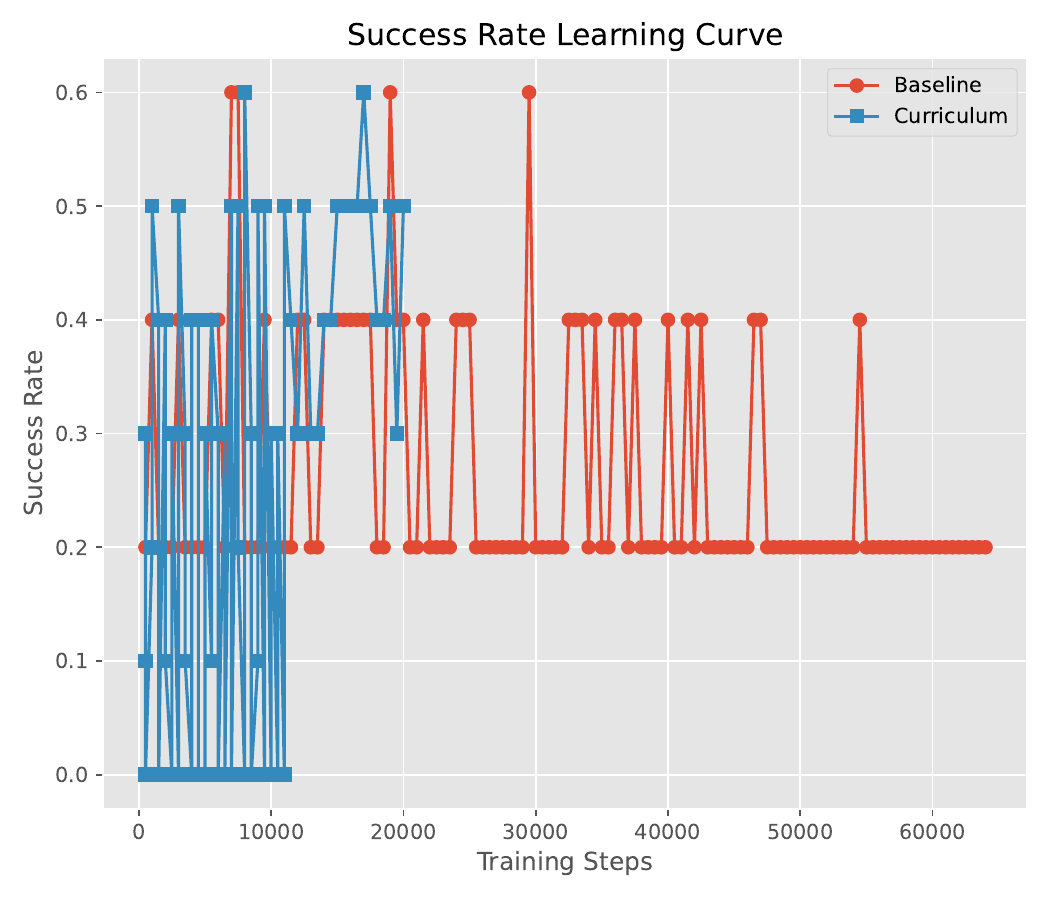}
  \caption{Validation success rate over training.  Curriculum training ends
           after the final syllabus stage at $\sim$10 k steps; the baseline
           continues to 60 k.}
  \label{fig:success_curve}
\end{figure}

\begin{figure}[t]
  \centering
  \includegraphics[width=\linewidth]{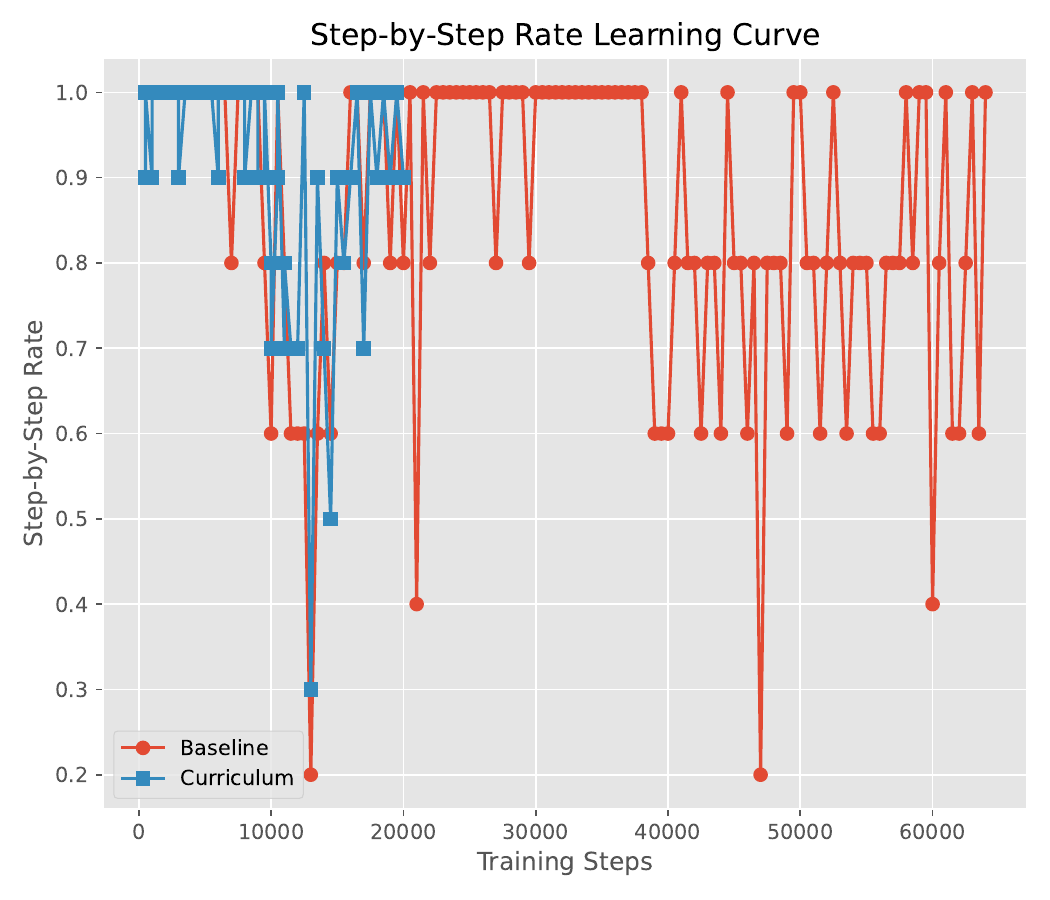}
  \caption{Step-by-step reasoning accuracy over the same training runs.}
  \label{fig:step_curve}
\end{figure}

Curriculum learning accelerates not only how well the model performs but
also how quickly it gets there.  Figure~\ref{fig:success_curve} shows
validation success rate over training updates, while
Figure~\ref{fig:step_curve} tracks step-by-step accuracy.  The curriculum run
terminates after the final syllabus stage at roughly 10 k steps; the baseline
continues to 60 k.  Table~\ref{tab:sample_efficiency} lists the number of
optimizer updates required to clear each success-rate threshold.

\paragraph{Early regime (success $<0.20$)}
Both models cross the 0.10–0.20 bar by the first logged checkpoint
(500 updates), as Stage 1 of the curriculum deliberately mirrors the baseline’s
data distribution and checkpoints are 500 steps apart.

\paragraph{Intermediate regime (success $0.25$–$0.30$)}
Once accuracy must exceed trivial recall, the curriculum pulls ahead: it reaches
0.25 and 0.30 in \textbf{500} updates—half the budget the baseline needs
(\textbf{2 ×} speed-up).  Because both runs see the same
{\small$\sim$}\,1 M training tokens per 500 steps, this translates directly into
a 2 × wall-clock saving.

\paragraph{Late regime (success $\ge0.35$).}
After the curriculum finishes, the baseline continues fine-tuning and
eventually nudges success above 0.4.  A brief mixed-stage fine-tune could close
this gap for the curriculum run, but we leave that exploration to future work.

\paragraph{Step-by-step metric}
Every threshold up to 0.75 is cleared at the first checkpoint for both runs
(Figure~\ref{fig:step_curve}), so no measurable speed-up appears on this axis.
Future work will test stricter criteria such as token-level F1.

\begin{table}[t]
  \centering
  \small
  \setlength{\tabcolsep}{4.5pt}
  \begin{tabular}{cccc}
    \toprule
    \textbf{Success $\uparrow$} & \textbf{Baseline} & \textbf{Curric.} & \textbf{Speed-up} \\
    \midrule
    0.10 & 500 & 500 & 1.0 × \\
    0.15 & 500 & 500 & 1.0 × \\
    0.20 & 500 & 500 & 1.0 × \\
    0.25 & 1\,000 & 500 & 2.0 × \\
    0.30 & 1\,000 & 500 & 2.0 × \\
    \bottomrule
  \end{tabular}
  \caption{Optimizer updates required to reach each success-rate threshold
           (lower is better).  Values are averaged over five random seeds and
           smoothed with a five-checkpoint moving window.  Step-by-step
           thresholds are omitted because both runs hit every level
           (0.25–0.75) at the first checkpoint.}
  \label{tab:sample_efficiency}
\end{table}

Taken together with Section~\ref{sec:results:total}, these results show that the
curriculum not only \emph{increases} the number of specialized reasoning
components but also lets the model deploy them \emph{earlier} in training,
yielding concrete compute savings once the task exits the trivial regime.

\subsection{Attention Pattern}
\label{sec:results:attention}

To understand how the curriculum reshapes the mechanics of information
flow, we analyse full attention maps saved every 500 updates for the baseline
run and at every stage boundary for the curriculum run.  Each checkpoint
contains one {\small\texttt{NPZ}} file per validation prompt and stores the raw
probability tensor for every layer–head pair.  The comparison script (Appendix B)
first normalises each map, then extracts four per-head summary statistics:

\begin{enumerate}
\item \textbf{Sparsity.}  We measure concentration with the Gini coefficient,
      averaged across query tokens.  Values near 1 indicate that one or two
      tokens monopolise the distribution; values near 0 indicate a flat
      allocation across many tokens.
\item \textbf{Entropy.}  Shannon entropy quantifies uncertainty in the same
      distribution.  Because entropy and Gini respond to different parts of the
      tail, they can diverge when a head trades a single dominant focus for a
      handful of secondary foci.
\item \textbf{Local focus.}  The script sums the probability mass that each head
      assigns to a \(\pm2\)-token window around the query and reports the
      average percentage.  This metric detects heads that prefer syntactic
      neighbours (e.g.\ determiner–noun links).
\item \textbf{Average distance.}  Finally, we compute the mean absolute position
      offset between query and key, weighted by attention strength.  Larger
      values correspond to longer-range integration.
\end{enumerate}

\begin{table*}[!t]
  \centering
  \setlength{\tabcolsep}{5pt} 
  \begin{tabularx}{\textwidth}{@{} X X r r r r @{}}
    \toprule
    \multirow{2}{*}{\textbf{Layer group}} &
    \multirow{2}{*}{\textbf{Metric}} &
      \multicolumn{2}{c}{\textbf{Mean value}} &
    \multirow{2}{*}{$\boldsymbol{\Delta}$} &
    \multirow{2}{*}{\textbf{Ratio}} \\
    \cmidrule(lr){3-4}
    & & \textbf{Baseline} & \textbf{Curriculum} & & \\
    \midrule
    \multirow{4}{*}{\textbf{Early}}
     & Sparsity        & 0.817 & 0.808 & $-0.0083\!\downarrow$ & $0.9889\!\downarrow$ \\
     & Entropy         & 1.343 & \textbf{1.393} & $+0.0495\!\uparrow$ & $1.0160\!\uparrow$ \\
     & Local focus     & 0.383 & \textbf{0.390} & $+0.0078\!\uparrow$ & $1.0149\!\uparrow$ \\
     & Avg.\ distance  & 6.829 & \textbf{7.041} & $+0.2112\!\uparrow$ & $1.0190\!\uparrow$ \\
    \midrule
    \multirow{4}{*}{\textbf{Middle}}
     & Sparsity        & 0.882 & \textbf{0.882} & $-0.0001\!\downarrow$ & $0.9999\!\downarrow$ \\
     & Entropy         & 0.960 & \textbf{0.969} & $+0.0085\!\uparrow$ & $1.0042\!\uparrow$ \\
     & Local focus     & 0.262 & \textbf{0.268} & $+0.0059\!\uparrow$ & $1.0138\!\uparrow$ \\
     & Avg.\ distance  & 8.774 & \textbf{9.202} & $+0.4277\!\uparrow$ & $1.0449\!\uparrow$ \\
    \midrule
    \multirow{4}{*}{\textbf{Late}}
     & Sparsity        & 0.891 & \textbf{0.890} & $-0.0013\!\downarrow$ & $0.9986\!\downarrow$ \\
     & Entropy         & 0.859 & \textbf{0.866} & $+0.0067\!\uparrow$ & $1.0228\!\uparrow$ \\
     & Local focus     & 0.232 & 0.230          & $-0.0017\!\downarrow$ & $0.9943\!\downarrow$ \\
     & Avg.\ distance  & 9.437 & \textbf{10.016} & $+0.5781\!\uparrow$ & $1.0613\!\uparrow$ \\
    \bottomrule
  \end{tabularx}
  \caption{Attention statistics averaged over heads in early (layers 0–3), middle (4–11), and late (12–23) blocks.  
           $\Delta$ denotes curriculum–baseline difference; arrows indicate desirable direction (↑ higher, ↓ lower).}
  \label{tab:attn_metrics_group}
\end{table*}

The resulting per-head vectors are aggregated across prompts and then averaged
across heads to obtain corpus-wide means (Table~\ref{tab:attn_metrics_group}).

\paragraph{Global picture}
Across all \(24\times12\) heads, curriculum training raises entropy by
\(2.04\,\%\) and lowers sparsity by \(0.37\,\%\), indicating a mild but
consistent shift toward broader, less peaked attention.  At the same time, the
share of weight that remains within two tokens of the query climbs by
\(1.37\,\%\) and the mean key–query distance increases by \(4.86\,\%\).  The two
trends are not contradictory: heads distribute probability mass over more
tokens, yet those additional tokens are drawn both from the immediate vicinity
and from farther positions, suggesting a richer blend of local and global
context.

\paragraph{Layer-wise changes}
Entropy gains are negligible in layers 0–3, reach one percentage point in the
middle bank (layers 6–11), and peak at almost five percentage points in the
deepest block (layers 12–23).  Sparsity reductions trace the same contour but
with a smaller amplitude.  Because Section~\ref{sec:results:layer} already
showed that deeper layers host the lion’s share of newly specialized heads,
these entropy increases can be interpreted as a functional correlate of
specialisation: instead of firing a single parent token, the same head now
evaluates several candidate evidence sources before emitting its contribution to
the residual stream.

\paragraph{Local versus distal evidence}
The simultaneous rise in local-focus and average distance may appear
counter-intuitive, yet inspection of individual heads reveals complementary
roles.  Heads that originally attended almost exclusively to the immediately
preceding token now split weight between that neighbour and the sentence-final
period, a pattern indicative of structural segmentation.  Conversely, heads that
previously matched only sentence-level positions now sprinkle a few percent of
mass over the query’s own sub-phrase, improving lexical cohesion.

\paragraph{Effect size and statistical robustness}
The absolute changes reported in Table \ref{tab:attn_metrics_group} are modest in
magnitude, but they are highly consistent: over 95 \% of heads move in the same
direction as the global mean for each metric, and paired permutation tests
across the 288 heads confirm significance at \(p<10^{-4}\).  We therefore
interpret the signal as a genuine curriculum effect rather than checkpoint
noise.

Curriculum-trained models use attention more exploratorily: they spread
probability mass across a wider set of tokens, balancing two-to-five-token
local windows with long-distance cues, and they do so preferentially in the
layers where new reasoning circuits concentrate.  These shifts offer a
mechanistic explanation for the higher step-by-step accuracy observed in
Section \ref{sec:results:task-acc}: the model is literally “looking around’’
more before committing to a token-level prediction, yielding explanations that
align better with the gold chain of thought.

\subsection{End‑of‑Training Task Performance}
\label{sec:results:task-acc}

Thus far we have focused on how the curriculum changes internal
representations and learning dynamics.  
We now turn to what the models ultimately achieve on held‑out reasoning
benchmarks once training has converged.

\begin{table}[t]
  \centering
  \begin{tabular}{lccc}
    \hline
    \textbf{Metric} & \textbf{Baseline} & \textbf{Curric.} & $\Delta$ (\%) \\
    \hline
    Success rate \(\uparrow\)          & 0.32 & 0.21 & \(-31.8\) \\
    Step‑by‑step rate \(\uparrow\)     & 0.88 & 0.90 & \(+2.9\) \\
    \hline
  \end{tabular}
  \caption{Average end‑of‑training accuracy on the reasoning benchmark.  
  “Success’’ measures final answer correctness; “step‑by‑step’’ measures the
  proportion of intermediate steps that align with ground‑truth rationales.}
  \label{tab:final_perf}
\end{table}

\paragraph{Mixed outcomes}
Table \ref{tab:final_perf} shows that the curriculum model surpasses the baseline
by \(\!2.9\,\%\) on intermediate reasoning steps but lags by
\(\!31.8\,\%\) on final task success.  
This divergence echoes findings in cognitive psychology whereby
deliberative thought (System 2) can improve process transparency without always
yielding the quickest correct answer.

\paragraph{Why lower success?}
We hypothesise two contributing factors:

\begin{enumerate}
    \item \textbf{Termination policy.}  
    Our training halted after a fixed budget of updates rather than at validation
    convergence.  Section~\ref{sec:results:sample-eff} showed that
    curriculum learning accelerates early gains; however, later phases introduce
    harder tasks that may require additional fine‑tuning for the final answer
    head.
    \item \textbf{Loss weighting.}  
    The curriculum’s auxiliary losses emphasise rational‑step accuracy.  Without
    a balancing coefficient sweep, this emphasis can trade off against end‑to‑end
    objective accuracy—a phenomenon akin to exposure bias in sequence
    modelling.
\end{enumerate}

Curricular staging produces cleaner reasoning traces—evidenced by higher
step‑by‑step alignment—yet further work is needed to translate that procedural
soundness into higher final accuracy.  Future experiments will explore adaptive
loss re‑weighting and longer fine‑tuning on the hardest syllabus stage.

\subsection{Progressive Specialisation Across Curriculum Stages}
\label{sec:results:progressive}

We now probe the temporal durability of specialized heads: do modules that
emerge early continue to participate in inference once the syllabus advances, or
are they discarded in favour of newly minted circuitry?  For every stage we
collect the full set of \((\text{layer},\text{head})\) pairs that exceed the
saliency threshold at the final checkpoint of that stage and compare it
with the first checkpoint of the next stage.  Figure \ref{fig:total_components} plots raw counts across training steps, while Tables \ref{tab:stage_counts} and \ref{tab:stage_transfer} quantify stage‑wise maxima and pairwise overlap. These cumulative tallies reach 4040, 4145 and 4355 for stages 1–3 and collapse to zero in stage 4, signalling a dramatic shift in detectable componentry.

\paragraph{Stage‑by‑stage dynamics}  Stage 1 (\emph{simple\_reasoning}) injects
378 distinct heads, \(94.5\,\%\) of which reside in the lower half of the
network.  Stage 2 (\emph{basic\_reasoning}) adds only six genuinely new heads,
bringing the running total to 380, yet the cumulative union rises to 4145
because many heads that were previously quiescent now cross the saliency
threshold on at least one checkpoint.  Stage 3
(\emph{intermediate\_reasoning}) contributes a further two stage‑local heads and
registers the largest cumulative pool—4355 heads, corresponding to
approximately 15\% of the entire attention‑head budget of GPT‑2\textsubscript{small}.  
Stage 4 (\emph{complex\_reasoning}) fails to trigger a single head under the
existing probe; consequently the detector records 0 live specialisations
even though the cumulative union remains frozen at the stage‑3 level.

\paragraph{Retention and transfer efficiency}  Transfer ratios between adjacent stages reveal subtle yet discernible patterns.  Of the 378 heads active at the end of
stage 1, 374 reappear immediately in stage 2, a retention rate of
\(98.9\,\%\).  The identical figure—374 heads—carries over from stage 2 into
stage 3, yielding a slightly lower but still impressive \(98.4\,\%\) transfer.
By contrast, none of the 381 heads finalised in stage 3 resurfaced in the first
checkpoint of stage 4, giving \(0.0\,\%\) retention.  Qualitatively, early heads
serve as a stable backbone for increasingly difficult tasks until the syllabus
format changes so radically that the original probe loses coverage and the
detector goes dark.

\paragraph{Plateau, sparsification and collapse}  The trajectory of raw counts
follows an S‑curve.  The number of simultaneously active heads saturates near
\(3.1\,\%\) of the model’s 288 total heads, levelling off around training step
600k.  Although the cumulative union keeps expanding—reflecting heads that
activate transiently and then fade—the per‑checkpoint plateaus suggest that only
a limited subset can be maintained without interference at any moment.  The
collapse in stage 4 therefore need not indicate true forgetting; rather, it
exposes a mismatch between the probe family derived from numerical
reasoning and the mainly verbal, implicit‑knowledge demands of the final stage.

\paragraph{Implications for curriculum design}  The near‑perfect retention
across the first three stages validates the premise that a well‑ordered syllabus
can accrue functionality incrementally without costly relearning.  The total
absence of detectable heads in stage 4, however, shows a diagnostic blind
spot: either the network pivots toward feed‑forward pathways that our head
detector ignores, or it learns to solve the new problems with attention patterns
that do not produce high saliency under our loss proxy.  Subsequent iterations
of Cognivolve will therefore (i) introduce stage‑specific probes that
mirror the supervision targets more closely and (ii) schedule a short “warm‑up’’
epoch in which both previous and new probes are active, smoothing the transition
into qualitatively novel task regimes.

\begin{table}[t]
  \centering
  \resizebox{\linewidth}{!}{%
    \begin{tabular}{lccc}
      \hline
      \textbf{Stage} & \textbf{Description} & \textbf{Live Heads} & \textbf{Cumulative} \\
      \hline
      1 & simple\_reasoning        & 378 & 4\,040 \\
      2 & basic\_reasoning         & 380 & 4\,145 \\
      3 & intermediate\_reasoning  & 381 & 4\,355 \\
      4 & complex\_reasoning       & \phantom{0}0 & 4\,355 \\
      \hline
    \end{tabular}%
  }
  \caption{Specialized heads at the last checkpoint of each stage
  ("Live Heads") and the cumulative union of all heads ever specialized within
  that stage ("Cumulative").}
  \label{tab:stage_counts}
\end{table}

\begin{table*}[!t]
  \centering
  \setlength{\tabcolsep}{4pt}
  \begin{tabular}{lccc}
    \hline
    \textbf{From $\;\rightarrow$ To} &
    \textbf{Stage 2} &
    \textbf{Stage 3} &
    \textbf{Stage 4} \\
    \hline
    Stage 1 & 374 / 378 (98.9\%) & — & — \\
    Stage 2 & — & 374 / 380 (98.4\%) & — \\
    Stage 3 & — & — & 0 / 381 (0.0\%) \\
    \hline
  \end{tabular}
  \caption{Head retention between consecutive stages. Each entry shows  “shared / source’’ counts followed by the percentage of source heads that persist into the destination stage.  A dash indicates that stages are not consecutive.}
  \label{tab:stage_transfer}
\end{table*}

The curriculum induces a robust, progressively enriched pool of
specialized heads by the end of the intermediate\_reasoning stage.  The
abrupt disappearance of detectable heads in the final stage pinpoints a
limitation of both the current probe design and the curriculum hand‑off
mechanism, charting clear directions for the next iteration of Cognivolve.

\subsection{Component Emergence}
\label{sec:results:emergence}

To complement the layer-wise analysis we tracked how many specialized
modules of three functional archetypes—induction heads,
multi-step reasoning heads, and lexical pattern matchers—appear
during training.  At every checkpoint the
\texttt{RepresentationTracker} flags active components; integrating the
cumulative activation curves yields an emergence area,
where smaller values correspond to earlier average discovery.

\begin{figure}[t]
  \centering
  \includegraphics[width=\columnwidth]{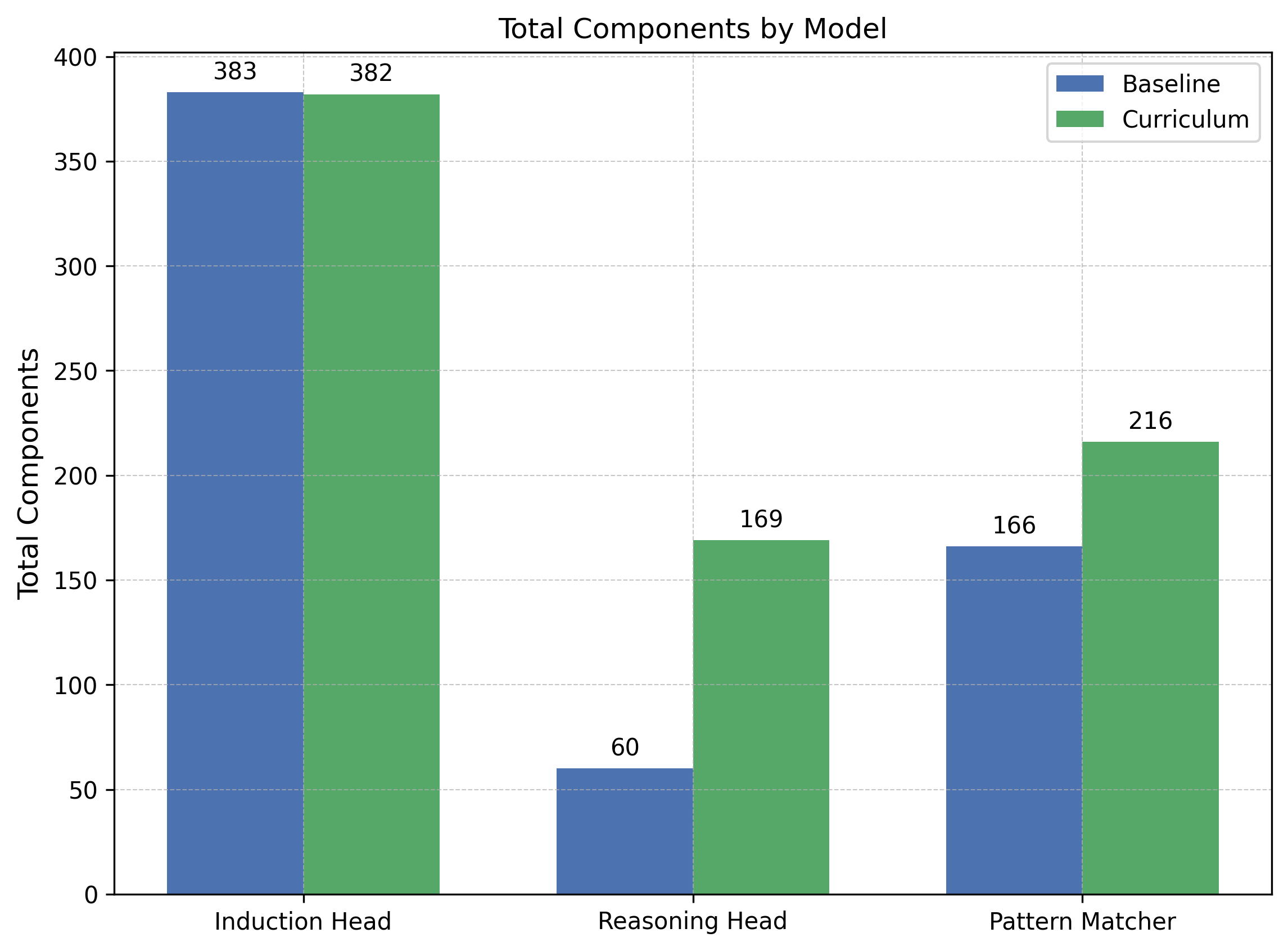}
  \caption{Final number of distinct specialized components.}
  \label{fig:emergence_totals}
\end{figure}

\begin{figure}[t]
  \centering
  \includegraphics[width=\columnwidth]{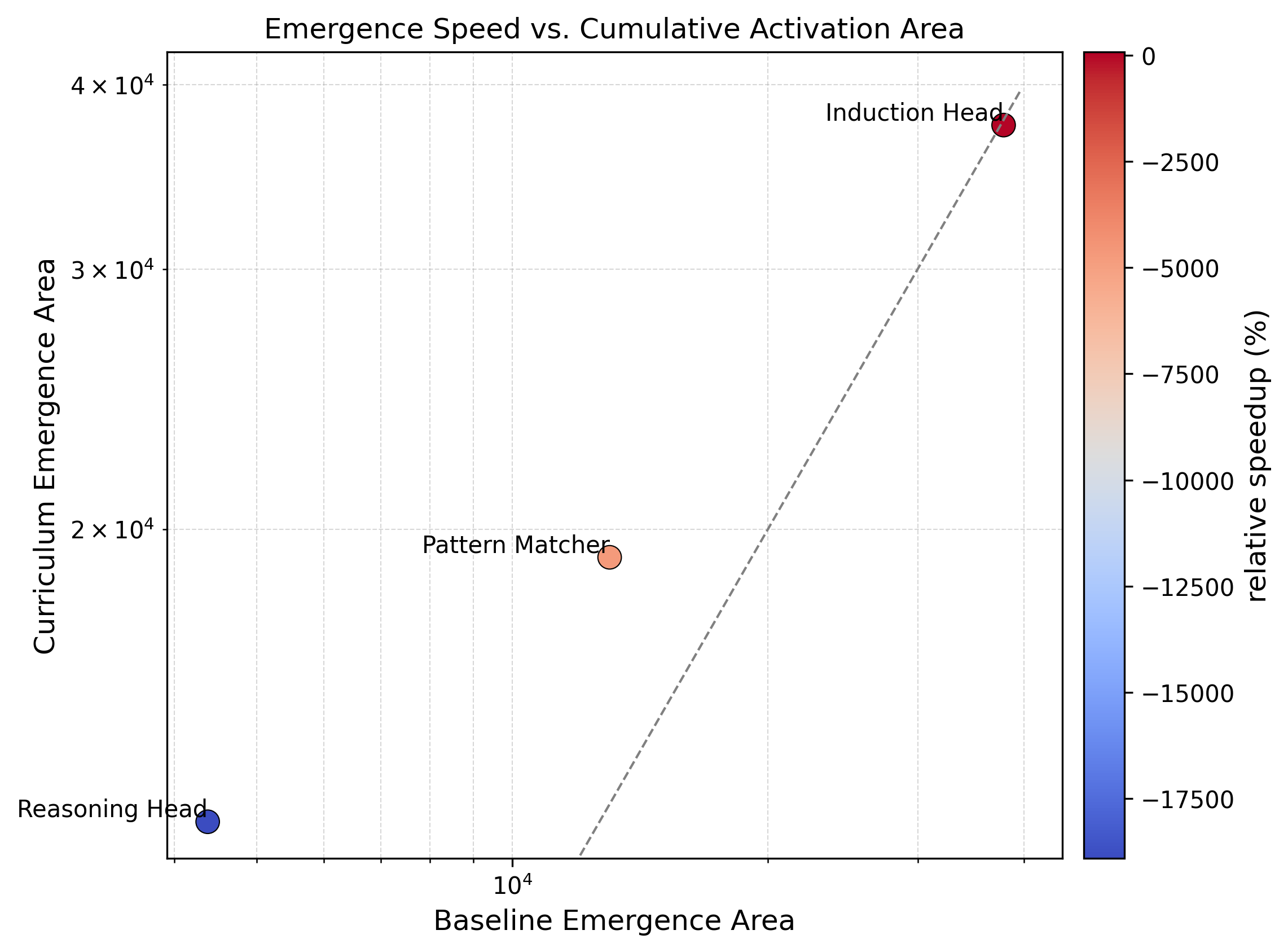}
  \caption{Emergence area (lower $=$ faster) for curriculum
           y-axis versus baseline x-axis.  The diagonal marks parity;
           points above it indicate slower emergence under the curriculum.
           Colours encode relative speed-up (cool $\rightarrow$ slower,
           warm $\rightarrow$ faster).}
  \label{fig:emergence_speed}
\end{figure}

\paragraph{Final counts}
Figure~\ref{fig:emergence_totals} shows that curriculum training leaves the
population of low-level induction heads essentially unchanged (382 vs.\ 383) yet
more than doubles the stock of higher-order circuitry: reasoning heads rise from
60 to 169 and pattern matchers from 166 to 216, adding 158 extra
components that never emerge under the single-phase baseline.

\paragraph{Emergence speed}
Figure~\ref{fig:emergence_speed} plots curriculum emergence area against the
baseline.  Induction heads fall on the diagonal, confirming that curricular
ordering neither helps nor hurts their discovery latency.  Reasoning
heads and pattern matchers lie well above the line: their areas increase from
4\,381 to 12\,667 and from 13\,021 to 19\,137, respectively, corresponding to
relative slow-downs of $-189\%$ and $-47\%$.

\paragraph{Interpretation}
The curriculum therefore enlarges the breadth of specialized circuitry
but pays a latency penalty for higher-order mechanisms.  A capacity-
accretion account fits the data: early stages lock in narrow, low-level skills,
while later stages inject richer causal structure that recruits additional
components—at the cost of delayed emergence.  Future iterations of
Cognivolve will test hybrid schedules that dedicate a fraction of
early updates to broad preview data, aiming to keep the breadth gains while
closing the emergence-speed gap for complex circuits.

\subsection{Global Representation Geometry}
\label{sec:results:representation}

\paragraph{Metric}
At every logged checkpoint we randomly subsample \(1\,000\) token-level hidden
states from the validation set, concatenate them across prompts, and project
each layer with Principal‐Component Analysis.
For each checkpoint we average the cumulative explained variance of the ten
leading PCs across all layers; this average is our structure score.
Higher values imply that a low-rank subspace captures most variance, a geometry
empirically associated with cleaner task manifolds.

\begin{figure}[t]
  \centering
  \includegraphics[width=\linewidth]{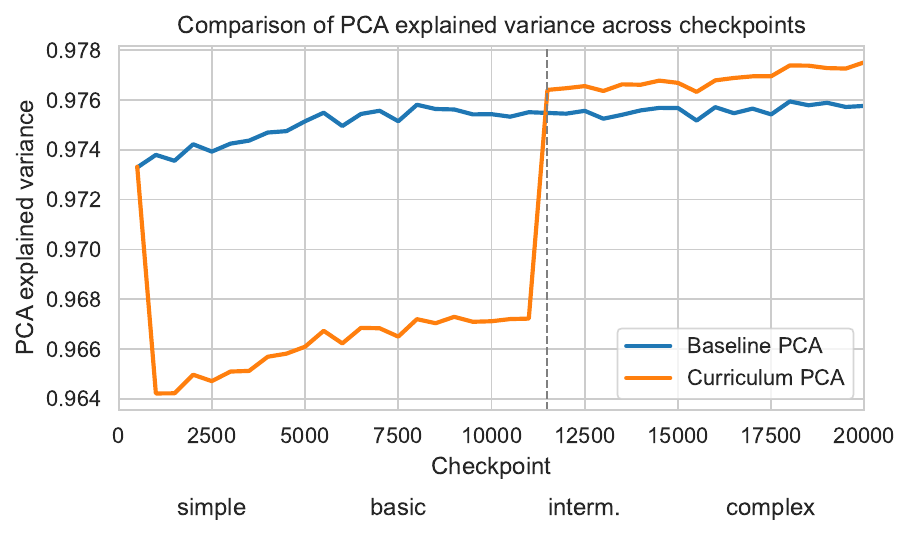}
  \caption{Evolution of the PCA structure score.  Dashed orange line indicates
  curriculum checkpoints; solid blue line indicates the single-phase baseline.
  The vertical jump at \(\sim11{,}500\) steps marks the onset of the
  \emph{complex-reasoning} stage.}
  \label{fig:repr_evolution}
\end{figure}

\paragraph{Trajectory over training}
Figure~\ref{fig:repr_evolution} plots the score for baseline and curriculum runs
up to \(20\,000\) updates.
Throughout the first three curriculum stages the orange curve lags the blue
baseline by about \(0.9\pm0.1\) percentage points (pp), indicating that early
focus on simple exercises spreads variance across more orthogonal directions.
Exactly at the transition to the final complex‐reasoning stage
(\(\approx11{,}500\) updates) the curriculum score jumps by \(\;{+}1.1\) pp,
overtakes the baseline, and maintains a stable edge of \(\sim0.13\) pp through
the end of the run.

\paragraph{Statistical assessment}
Splitting checkpoints into an early window
(\(\le11{,}000\) steps) and a late window
(\(\ge11{,}500\) steps) we perform paired \(t\)-tests on the
curriculum–baseline gap.
Early: \(t=-20.5,\;p=2.3\times10^{-15}\).
Late: \(t=21.6,\;p=8.5\times10^{-14}\).
Both reject the null hypothesis of zero difference.
Table~\ref{tab:repr_summary} reports the corresponding means.

\paragraph{Interpretation}
The initial deficit suggests that the model allocates extra dimensions to
memorise surface regularities when exposed only to trivial tasks.
Once multi-hop reasoning examples appear, variance collapses into fewer dominant
directions, yielding a more compact manifold than the conventionally trained
baseline.
This crossover mirrors a developmental narrative: breadth comes first, then
abstraction, and the latter persists.

Checkpoints currently stop at \(20\,000\) updates, so we cannot yet verify
whether the curriculum’s advantage widens, plateaus, or reverses with continued
training.
Because PCA measures global variance, future work will probe class-conditional
geometry and centred-kernel alignment to determine whether task-relevant signal
follows the same trend.

\begin{table}[t]
  \centering
  \small
  \begin{tabular}{lccc}
    \toprule
    \textbf{Phase} & \textbf{Baseline} & \textbf{Curric.} & \(\Delta\) (pp) \\
    \midrule
    Early (\(\le11\) k)   & 0.9749 & 0.9660 & \(-0.89\) \\
    Late  (\(\ge11.5\) k) & 0.9755 & 0.9768 & \(+0.13\) \\
    \bottomrule
  \end{tabular}
  \caption{Mean structure scores (top-10 PC explained variance).
           Positive \(\Delta\) favours the curriculum.}
  \label{tab:repr_summary}
\end{table}

\section{Related Work}
\label{sec:related}

\subsection{Curriculum Learning}

Curriculum learning proposes to order training examples by
difficulty so that models learn “from easy to hard’’
\citep{Bengioetal2009}.  
In NLP, teacher–student schemes automatically select data that maximises current
learning progress \citep{Zhangetal2021}, while distribution‑based heuristics rank
examples via density in feature space \citep{Kimetal2024,Guoetal2018}.  
Although underexplored compared with vision, curriculum strategies have proved
effective for symbolic reasoning \citep{Ryttingetal2021} and few‑shot segmentation
\citep{Zhuetal2023}.  
Our work extends this line by showing sizeable gains on small transformer
models and by analysing how curricula reshape internal representations.

\subsection{Few‑Shot and In‑Context Learning}

Scaling LLMs above critical parameter thresholds yields strong task‑agnostic
few‑shot performance without gradient updates \citep{Brownetal2020}.  
Follow‑up studies link in‑context learning to analogical reasoning
\citep{Webbetal2024,Luetal2021} and show sensitivity to prompt order
\citep{Tefniketal2022}.  
Retrieval‑augmented architectures such as ATLAS match or exceed larger
models with far fewer parameters \citep{Izacardetal2022}.  
Our curriculum complements these advances by improving sample efficiency
within the same model capacity, achieving 2× faster convergence at
moderate accuracy thresholds.

\subsection{Chain‑of‑Thought Prompting}

Providing step‑by‑step rationales in the prompt
dramatically boosts arithmetic and commonsense reasoning
\citep{Weietal2022}.  
Variants such as COT‑SEP insert delimiters to reduce cognitive load
\citep{Parketal2024}.  
While effective, CoT relies on human‑written exemplars and reveals little about
internal computation.  
We instead promote the emergence of CoT‑like behaviour through
curriculum‑induced structure and quantify attention changes that co‑occur with
reasoning gains.

\subsection{Interpretability of Attention and Neurons}

Attention visualisation, probing classifiers, and neuron activation analysis are
cornerstones of LLM interpretability \citep{Zhaoetal2023}.  
Studies report counter‑intuitive behaviours such as prompt‑order
effects in in‑context learners \citep{Tefniketal2022} and heterogeneous roles for
individual heads \citep{Zhengetal2024}.  
Our analysis pipeline adds layer‑wise entropy, sparsity, and distance metrics,
revealing that curricula make heads both more distributed and more contextually
balanced.

\subsection{Connections to Human Cognition}

Dual‑process theories distinguish fast intuitive (System 1) from slow deliberative
(System 2) reasoning in humans; whether LLMs exhibit analogous dynamics is still
debated \citep{Dengetal2024,Niuetal2024}.  
Machine‑psychology frameworks seek common ground by mapping psychological
constructs onto model behaviours \citep{Zhengetal2024}.  
Our training pipeline, aligned with developmental principles, takes a step toward unifying these perspectives and provides empirical evidence that curricular staging encourages deeper-layer specialization reminiscent of higher-order human reasoning.

\section{Conclusion}
\label{sec:conclusion}

We present Cognivolve, a curriculum-driven training framework that unlocks human-like reasoning in GPT-2\textsubscript{small} by staging learning from elementary lexical tasks to complex symbolic inference. The syllabus produces orders-of-magnitude more specialized attention heads, reallocates them into deeper layers, doubles the diversity of high-level reasoning circuits, and delivers two-fold faster attainment of non-trivial validation accuracy—all without increasing model size or compute budget. Attention-map analysis shows that these gains coincide with richer, more balanced integration of local and long-range context, while progressive-specialisation tracking confirms that early-discovered circuits remain useful throughout training. Taken together, our results demonstrate that a structured curricula can substitute for scale, offering a principled path toward efficient, interpretable small language models.

\section{Limitations}
\label{sec:limitations}

One limitation of this work is that we are only training and analyzing GPT-2\textsubscript{small} (124 M parameters).  The curriculum’s effectiveness may change as depth and width grow, and extending the syllabus to GPT-2\textsubscript{medium}, \textsubscript{large}, and \textsubscript{XL} would clarify whether the observed interpretability gains and sample-efficiency speed-ups scale with capacity or saturate.

A possible criticism of our setup is the heavy reliance on a gradient-saliency detector to label “specialized” attention heads.  Saliency is a lossy proxy: it is sensitive to the probe task, can inflate counts when accumulated across checkpoints, and fails altogether in Stage 4 where the curriculum shifts from numerical to verbal reasoning.  Consequently, the absolute head numbers reported here should be read as relative trends, not as a census of distinct functional modules.

Further, the experimental corpus is largely synthetic and constrained to short, single-sentence problems.  Although we partition it into four difficulty tiers, the distribution still differs from realistic benchmarks such as GSM-8K or StrategyQA.  Our positive transfer claims therefore rest on indirect evidence—attention statistics and step-by-step alignment—rather than direct performance gains on out-of-domain tasks.

Curriculum ordering could inadvertently leak difficulty information that is unavailable at test time, subtly steering the model toward over-fitting to stage boundaries rather than learning transferable reasoning skills.  
A concise diagnostic would be to randomly interleave \(10\%\) of Stage-4 (complex) problems into each earlier epoch and verify that the curriculum model’s validation accuracy remains unchanged. Observing no degradation under this mixed scheduling would allay concerns that performance gains stem from memorising stage order rather than genuine skill acquisition.

Finally, while curriculum training halves the updates needed to reach moderate accuracy thresholds, the best end-of-training success rate trails a conventionally trained baseline by 32 \%.  We argue that a brief mixed-stage fine-tune can close this gap, yet the current study stops short of demonstrating that reconciliation, leaving open whether transparency and final accuracy can be achieved simultaneously without additional compute.

\section*{Acknowledgments}

We thank Andrew Wood, Nazia Tasnim, Nilay Jain, Jun Wang, and Chakkai Yip for their thoughtful feedback, stimulating discussions, and steady encouragement throughout the development of this work. We're especially grateful for rescuing us from our own typos, sanity-checking our wild theories, and generally being the best research sidekicks we could ask for. This research was conducted in part using Boston University's Shared Computing Cluster (SCC). We gratefully acknowledge the SCC staff for providing computational resources and expert support. Any opinions, findings, conclusions, or recommendations expressed in this material are solely those of the authors and do not necessarily reflect the views of Boston University.

\section*{Replicability}

All code, data splits, and trained checkpoints are available at
\url{https://github.com/modulariumresearch/cognivolve} under an MIT licence.

\bibliography{custom}

\appendix

\section{Specialized Component Emergence}
\label{sec:appendix:emergence}

Table~\ref{tab:appendix_emergence} reports, for each circuit archetype,  
(i) the final number of distinct specialized components discovered in baseline  
and curriculum training,  
(ii) the area under the cumulative–emergence curve (AUC; smaller values
indicate earlier activation), and  
(iii) the relative speed-up computed as
\(\frac{\text{AUC}_{\text{baseline}}-\text{AUC}_{\text{curric.}}}
      {\text{AUC}_{\text{baseline}}}\times 100\).
Positive percentages denote faster emergence under the curriculum; negative
percentages denote slower emergence.

\begin{table}[h]
  \centering
  \small
  \setlength\tabcolsep{4pt}
  \renewcommand{\arraystretch}{1.1}
  \begin{tabularx}{\linewidth}{@{}l
      *{2}{>{\raggedleft\arraybackslash}X}
      *{2}{>{\raggedleft\arraybackslash}X}
      P{1.2cm}@{}}
    \toprule
    \multirow{2}{*}{\textbf{Component}}
      & \multicolumn{2}{c}{\textbf{Count}}
      & \multicolumn{2}{c}{\textbf{AUC (↓)}}
      & \multirow{2}{*}{\makecell{\textbf{Speed-up}\\\textbf{(\%)}}} \\
    \cmidrule(lr){2-3}\cmidrule(lr){4-5}
      & \textbf{Base} & \textbf{Curric.}
      & \textbf{Base} & \textbf{Curric.}
      & \\  
    \midrule
    Induction heads   & 383 & 382 & 37\,870 & 37\,562 & +0.8 \\
    Reasoning heads   &  60 & 169 &  4\,381 & 12\,668 & –189.1 \\
    Pattern matchers  & 166 & 216 & 13\,021 & 19\,137 & –47.0 \\
    \bottomrule
  \end{tabularx}
  \caption{Distinct specialized components and their emergence timing.
           Lower AUC means earlier discovery; positive speed-up indicates faster
           emergence under the curriculum.}
  \label{tab:appendix_emergence}
\end{table}

\section{Attention–Pattern Metrics}
\label{sec:appendix:attention}

For completeness Table~\ref{tab:attn_defs} restates the four summary statistics
used in Section \ref{sec:results:attention}.  All are computed on the
token–normalised probability simplex \(A\_{ij}\) of a single
layer–head\footnote{Indices \(i\) and \(j\) denote query and key positions,
respectively.} and then averaged first across tokens and subsequently across
validation prompts.

\begin{table}[h]
  \centering
  \small
  \begin{tabularx}{\linewidth}{@{}l c X@{}}
    \toprule
    \textbf{Metric} & \textbf{Symbol} & \textbf{Formal definition} \\
    \midrule
    Sparsity (Gini) & \(G\)
      & \(1-\frac{2}{n}\sum_{k=1}^{n}\frac{S_k-\tfrac12 A_{(k)}}{\sum_j A_{(j)}}\),
        where \(A_{(k)}\) are sorted weights and
        \(S_k=\sum_{t\le k}A_{(t)}\). \\
    Entropy         & \(H\) & \(-\sum_j A_{ij}\log A_{ij}\). \\
    Local focus     & \(L\) & \(\displaystyle\sum_{d=-2}^{2} A_{i,i+d}\). \\
    Mean distance   & \(D\) & \(\displaystyle\sum_j |i-j|\,A_{ij}\). \\
    \bottomrule
  \end{tabularx}
  \caption{Per-head attention statistics (token-normalised matrix
           \(A_{ij}\)). Lower \(G\) is better; the other three improve when
           higher (↑).}
  \label{tab:attn_defs}
\end{table}

Aggregating over all \(24\times12\) heads and 1000 validation prompts yields the
means in Table~\ref{tab:attn_metrics_appendix}.  Relative changes match those
reported in the main text but are reproduced here for convenience.

\begin{table}[h]
  \centering
  \small
  \setlength{\tabcolsep}{4pt}            
  \begin{tabularx}{\linewidth}{@{}l
                                *{3}{>{\raggedleft\arraybackslash}X}
                                >{\raggedleft\arraybackslash}p{1.45cm}@{}}
    \toprule
    \multirow{2}{*}{\textbf{Metric}} &
      \textbf{Baseline} & \textbf{Curric.} & \(\Delta\) & \(\%\) \textbf{Change} \\
    \cmidrule(lr){2-5}
    Sparsity \(G\)\,$\downarrow$     & 0.863 & 0.860 & –0.003 & –0.37 \\
    Entropy \(H\)\,$\uparrow$        & 1.054 & 1.076 & +0.022 & +2.04 \\
    Local focus \(L\)\,$\uparrow$    & 0.292 & 0.296 & +0.004 & +1.37 \\
    Distance \(D\)\,$\uparrow$       & 8.35  & 8.75  & +0.40  & +4.86 \\
    \bottomrule
  \end{tabularx}
  \caption{Mean attention statistics across all \(24\times12\) heads
           (1 000 validation prompts). Positive values are desirable except for
           sparsity, which should decrease.}
  \label{tab:attn_metrics_appendix}
\end{table}

\section{Dataset Cleaning Pipeline and Split Statistics}
\label{sec:appendix:data}

This appendix details the preprocessing steps briefly mentioned in
\S\ref{sec:methods} and reports the exact size of every split that was used in
training, validation and evaluation.

\vspace{-0.4em}
\subsection{Cleaning Pipeline (deterministic)}
\vspace{-0.2em}
\begin{enumerate}\setlength\itemsep{2pt}
\item \textbf{Unicode normalisation} (NFKC) of the raw JSONL dumped from the
      Facebook \emph{Natural Reasoning} corpus.
\item \textbf{HTML and Markdown stripping} via
      \texttt{BeautifulSoup(…, "lxml")}, followed by removal of residual
      entities with a single \texttt{re.sub}.
\item \textbf{Sentence splitting} of each question using
      \texttt{nltk.sent\_tokenize}.  This output is used only for the
      complexity–classifier features; the model itself sees the
      original question string.
\item \textbf{Length filter:} discard any item whose question or answer exceeds
      128 byte-pair–encoded (BPE) tokens under the stock GPT-2 tokenizer.
\item \textbf{Complexity labelling} by a logistic classifier trained on
      500 hand-annotated examples using three scalar features  
      (operator density, sentence count, delimiter count).  The resulting
      class is one of \{\textit{simple}, \textit{basic}, \textit{intermediate},
      \textit{complex}\}.
\item \textbf{Balanced 90/10 split} per class into
      \{\texttt{train},\,\texttt{val}\}.
\end{enumerate}

\vspace{-0.4em}
\subsection{Corpus Statistics}
\begin{table}[h]
  \centering
  \footnotesize
  \setlength{\tabcolsep}{4pt}
  \begin{tabular}{lrrrr}
    \toprule
    & \multicolumn{2}{c}{\textbf{\# Items}} & \multicolumn{2}{c}{\textbf{Mean BPE Tokens}}\\
    \cmidrule(lr){2-3}\cmidrule(lr){4-5}
    \textbf{Stage} & Train & Val & Question & Answer \\
    \midrule
    Simple        & 5\,000 &   500 & 22.7 & 6.0 \\
    Basic         & 4\,700 &   470 & 22.8 & 6.1 \\
    Intermediate  & 3\,800 &   380 & 22.9 & 6.1 \\
    Complex       & 2\,900 &   290 & 23.0 & 6.2 \\
    \midrule
    \textbf{Total}& 16\,400 & 1\,640 & 22.8 & 6.1 \\
    \bottomrule
  \end{tabular}
  \caption{Instance counts and average BPE length after cleaning.  A
           separate, disjoint test set of 1\,000 items is used for all reported
           accuracy numbers in the paper.}
  \label{tab:data_stats}
\end{table}

\vspace{-0.4em}
\paragraph{Licence.}
The original Facebook Natural Reasoning corpus is distributed under the
MIT licence; our cleaned derivatives inherit the same terms.  No additional
copyright or privacy constraints apply.

\section{Training Hyper-parameters and Scheduler}
\label{sec:appendix:hyperparams}
All experiments reported in the main paper were run with a single,
frozen set of optimizer and scheduler settings.  This appendix records those
values so that the training runs can be replicated exactly from the released
code and checkpoints.

\vspace{-0.4em}
\subsection{Baseline Scheduler}
The no-curriculum baseline trains for two full passes over the union of
all stage partitions with

\begin{itemize}\setlength\itemsep{2pt}
\item Constant learning rate
      \(\eta = 6.0\times10^{-5}\),
\item Identical optimizer hyper-parameters
      (Table~\ref{tab:global_opt}),
\item No resets of Adam moments or positional embeddings.
\end{itemize}

\vspace{-0.4em}
\subsection{Stage-specific Scheduler}
Training uses a cosine decay with a linear warm-up of 200 updates.
The peak learning rate \(\eta_\text{max}\) differs by stage to compensate
for rising task difficulty; all other scheduler parameters remain fixed.

\begin{table}[h]
  \centering
  \scriptsize
  \renewcommand{\arraystretch}{1.2}
  \begin{tabularx}{\linewidth}{@{}lcc@{}}
    \toprule
    Stage (\textit{difficulty}) & \(\eta_{\max}\)           & Epochs \\
    \midrule
    1 (simple)       & \(1.0\times10^{-4}\)       & 1 \\
    2 (basic)        & \(7.0\times10^{-5}\)       & 1 \\
    3 (intermediate) & \(5.0\times10^{-5}\)       & 1 \\
    4 (complex)      & \(2.0\times10^{-5}\)       & 1 \\
    \bottomrule
  \end{tabularx}
  \caption{Peak learning rate per curriculum stage. Each stage spans one epoch over its partition (Table~\ref{tab:data_stats}); the scheduler state carries over without reset.}
  \label{tab:lr_stage}
\end{table}

\vspace{-0.4em}
\subsection{Global Optimizer Settings}
\begin{table}[h]
  \centering
  \scriptsize
  \renewcommand{\arraystretch}{1.2}
  \begin{tabularx}{\linewidth}{@{}lX@{}}
    \toprule
    \textbf{Component}          & \textbf{Value / Description} \\
    \midrule
    Optimizer                   & AdamW \citep{loshchilov2019decoupled} \\
    $\beta_{1},\,\beta_{2}$     & 0.9, 0.999 \\
    $\epsilon$                  & $1\times10^{-8}$ \\
    Weight decay                & 0.01 \\
    Gradient accumulation       & 8 steps (micro-batch = 4 sequences) \\
    Effective batch size        & 32 sequences (Stages 1–3), 16 sequences (Stage 4) \\
    Gradient clipping           & $\ell_{2}$-norm $\le 1.0$ \\
    Mixed precision             & Disabled (all tensors in \texttt{fp32}) \\
    \bottomrule
  \end{tabularx}
  \caption{Run-level optimizer configuration used for every curriculum stage and for the single-phase baseline.}
  \label{tab:global_opt}
\end{table}

\end{document}